\documentclass[10pt,twocolumn,letterpaper]{article}

\usepackage{wacv}
\usepackage{times}
\usepackage{epsfig}
\usepackage{graphicx}
\usepackage{amsmath}
\usepackage{amssymb}
\usepackage{algorithm}
\usepackage{algorithmic}
\usepackage{adjustbox}

\newtheorem{theorem}{Theorem}

\wacvfinalcopy % *** Uncomment this line for the final submission

\def\wacvPaperID{289} % *** Enter the wacv Paper ID here

\ifwacvfinal\fi
\begin{document}
%%%%%%%%% TITLE
\title{Learning From Less Data: A Unified Data Subset Selection and Active Learning Framework for Computer Vision}

% Authors at the same institution
%\author{First Author \hspace{2cm} Second Author \\
% Authors at different institutions
\author{Vishal Kaushal\\
IIT Bombay\\
Mumbai, India\\
{\tt\small vkaushal@cse.iitb.ac.in}
\and
\and
Rishabh Iyer\\
Microsoft\\
Redmond, Washington, USA\\
{\tt\small rishi@microsoft.com}
\and 
Suraj Kothawade \\
IIT Bombay\\
Mumbai, India\\
{\tt\small surajkothawade@cse.iitb.ac.in}
\and 
Rohan Mahadev \\
AITOE Labs\\
Mumbai, India\\
{\tt\small rohan@aitoelabs.com}
\and
Khoshrav Doctor\\
University of Massachusetts\\
Massachusetts, USA\\
{\tt\small kdoctor@cs.umass.edu}
\and
Ganesh Ramakrishnan\\
IIT Bombay\\
Mumbai, Maharashtra, India\\
{\tt\small ganesh@cse.iitb.ac.in}
}

\maketitle
\ifwacvfinal\thispagestyle{empty}\fi

%%%%%%%%% ABSTRACT
\begin{abstract}
   Supervised machine learning based state-of-the-art computer vision techniques are in general data hungry. Their data curation poses the challenges of expensive human labeling, inadequate computing resources and larger experiment turn around times. Training data subset selection and active learning techniques have been proposed as possible solutions to these challenges. A special class of subset selection functions naturally model notions of diversity, coverage and representation and can be used to eliminate redundancy thus lending themselves well for training data subset selection. They can also help improve the efficiency of active learning in further reducing human labeling efforts by selecting a subset of the examples obtained using the conventional uncertainty sampling based techniques. In this work, we empirically demonstrate the effectiveness of two diversity models, namely the Facility-Location and Dispersion models for training-data subset selection and reducing labeling effort. We demonstrate this across the board for a variety of computer vision tasks including Gender Recognition, Face Recognition, Scene Recognition, Object Detection and Object Recognition. Our results show that diversity based subset selection done in the right way can increase the accuracy by upto 5 - 10\% over existing baselines, particularly in settings in which less training data is available. This allows the training of complex machine learning models like Convolutional Neural Networks with much less training data and labeling costs while incurring minimal performance loss.
\end{abstract}

%%%%%%%%% BODY TEXT
\section{Introduction}

Deep Convolutional Neural Network based models are today the state-of-the-art for most computer vision tasks. Seeds of the idea of deep learning were sown around the late 90's ~\cite{lecun1998gradient} and it gained popularity in 2012 when AlexNet ~\cite{krizhevsky2012imagenet} won the challenging ILSVRC (ImageNet Large-Scale Visual Recognition Challenge) 2012 competition ~\cite{russakovsky2015imagenet}, demonstrating an astounding improvement over the then state-of-the-art image classification techniques. This was soon followed by an upsurge of deep models for computer vision tasks from all over the community. This renewed interest in CNNs is due to the accessibility of large training sets and increased computational power, thanks to GPUs. VGGNet ~\cite{simonyan2014very} demonstrated that a simpler, but deeper model can be used to improve accuracies. Then came the even deeper GoogLeNet ~\cite{szegedy2015going}, the first CNN to be fundamentally architecturally different from AlexNet. GoogLeNet introduced the idea that CNN layers didn't always have to be stacked up sequentially. ResNet ~\cite{he2016deep} was even deeper, a phenomenal 152 layer deep network with an incredible error rate of 3.57\%, beating humans in the image classification task. State-of-the-art face recognition techniques such as DeepFace ~\cite{taigman2014deepface}, DeepID3 ~\cite{sun2015deepid3}, Deep Face Recognition ~\cite{parkhi2015deep} and FaceNet ~\cite{schroff2015facenet} also consist of deep convolutional networks. Similar is the story with state-of-the-art scene recognition techniques ~\cite{zhou2014learning}, and techniques for other computer vision tasks such as age and gender classification~\cite{levi2015age} etc. Similarly, for object detection tasks, the first significant advancement in deep learning was made by RCNNs ~\cite{girshick2014rich}, which was followed by Fast RCNN ~\cite{girshick2015fast} and then Faster RCNN ~\cite{ren2015faster} toward performance improvement. More recently, YOLO ~\cite{redmon2016you} and YOLO9000 ~\cite{redmon2016yolo9000} have emerged as the state-of-the-art in object detection. YOLO's architecture is inspired by GoogLeNet and consists of 24 convolutional layers followed by two fully connected layers.\looseness-1

Every coin, however, has two sides and so is the case with deep learning. While deeper models are increasingly improving for computer vision tasks, they pose the following challenges: a) Increased training complexity and computational costs, b) Larger inference time, c) Larger experimental turn around times and difficulty in hyper-parameter tuning, and d) Higher costs and more time for labeling.

Training complexity and huge data requirements are owing to the depth of the network and the large number of parameters to be learnt. A large deep neural net with a large  number of parameters to learn (and hence a large degree of freedom) has a very complex and extremely non-convex error surface to traverse and thus it requires a great deal of data to successfully search for a reasonably optimal point on that surface. Each of the CNN architectures (AlexNet~\cite{krizhevsky2012imagenet}, ZFNet~\cite{zeiler2014visualizing}, VGGNet and GoogleNet were trained over a span of several days (and in even weeks in some cases) on a couple of GPUs. Training with several GPUs together can reduce the time taken, but this results in slow experimental turn around time especially for hyper-parameter tuning which is very important for getting these models to work in practice. 

Orthogonal to the challenge of training complexity is the challenge of unavailability of labeled data. Human labeling efforts are costly and grow exponentially with the size of the dataset ~\cite{vondrick2013efficiently}. A lot of data today comes from videos, which have a naturally associated redundancy. 

\subsection{Existing Work}
In this section, we review existing work addressing the problems of increased training complexity, larger turn around times, increased complexity for inference (runtime) and increased labeling costs.

\noindent \textbf{Network Architechture Modifications for reducing training/inference time: } One way researchers have addressed this challenge is through architectural modifications to the network. For example, by making significant architectural changes, GoogLeNet improved utilization of the computing resources inside the network thus allowing for increased depth and width of the network while keeping the computational budget constant. Similarly, by explicitly reformulating the layers as learnt residual functions with reference to the layer inputs, instead of learning unreferenced functions, ResNet allows for easy training of much deeper networks. Highway networks ~\cite{srivastava2015training} introduce a new architecture allowing a network to be trained directly through simple gradient descent. With the goal of reducing training complexity, some other studies ~\cite{ba2014deep, hinton2015distilling, iandola2016squeezenet} have also focused on model compression techniques. On similar lines, studies like ~\cite{levi2015age} propose simpler network architectures in favor of availability of limited training data. 

\noindent \textbf{Transfer Learning for reduced training/inference time: } Other approaches advocate use of pre-trained models, trained on large data sets and presented as `ready-to-use' but they rarely work out of the box for domain specific computer vision tasks (such as detecting security sensitive objects). This idea is called Transfer Learning~\cite{donahue2014decaf}. %It thus becomes important to create customized models, which again potentially require significant amount of training data. Domain adaptation techniques like transfer learning and fine tuning allow creation of customized models with much lesser data. 
Transfer learning allows models trained on one task to be easily reused for a different task ~\cite{pan2010survey}. Several studies ~\cite{donahue2014decaf, yosinski2014transferable, long2015learning} have analyzed the transferability of features in deep convolutional networks across different computer vision tasks. It has been demonstrated that transferring features even from distant tasks can be better than using random features.

\noindent \textbf{Reducing Labeling Costs: } Several approaches have been proposed to reduce labeling costs. In ~\cite{oquab2015object} the authors describe a weakly supervised convolutional neural network (CNN) for object classification that relies only on image-level labels, yet can learn from cluttered scenes containing multiple objects to produce their approximate locations. Zero-shot learning ~\cite{socher2013zero, changpinyo2016synthesized} and one-shot learning~\cite{vinyals2016matching} techniques also help address the issue of dearth of annotated examples. In ~\cite{gavves2015active}, Gavves et al combine techniques from transfer learning, active learning and zero-shot learning to reuse existing knowledge and reduce labeling efforts. Zero-shot learning techniques don't expect any annotated samples, unlike active learning and transfer learning that assume either the availability of at least a few labeled target training samples or an overlap with existing labels from other datasets. \cite{doersch2017multi} propose a technique of combining self supervised learning tasks (i.e. where training data can be collected without manual labeling). %By using a max-margin formulation (which transfers knowledge from zero-shot learning models used as priors) they present two basic conditions to choose greedily only the most relevant samples to be annotated, and this forms the basis of their active learning algorithm. 

\noindent \textbf{Active Learning in Computer Vision: } The core idea behind active learning~\cite{settles2010active} is that a machine learning algorithm can achieve greater accuracy with fewer training labels if it is allowed to choose the data from which it learns. There have been very few studies in active learning and subset selection for Computer Vision tasks. Using an information theoretic objective function such as mutual information between the labels, Sourati et al developed a framework ~\cite{sourati2016classification} for evaluating and maximizing this objective for batch mode active learning. Another recent study has adapted batch active learning to deep models ~\cite{ducoffescalable}. Most batch-active learning techniques involve the computation of the Fisher matrix which is intractable for deep networks. Their method relies on computationally tractable approximation of the Fisher matrix, thus making them relevant in the context of deep learning.

\noindent \textbf{Diversified Data Subset Selection and Active Learning: } A common approach to training data subset selection is to use the concept of a coreset ~\cite{agarwal2005geometric}, which aims to efficiently approximate various geometric extent measures over a large set of data instances via a small subset. Submodular functions naturally model the notion of diversity, representation and coverage and hence submodular function optimization has been applied to recommendation systems to recommend relevant and diverse items that explicitly account for the coverage of user interests~\cite{shaframework}.
Submodular Functions form natural models for training data subset selection~\cite{wei2015submodularity, wei2014submodular}. In particular, the data-likelihood functions for the Naive Bayes and Nearest Neighbor classifiers turn out to be Feature based and Facility Location functions respectively~\cite{wei2015submodularity}. Therefore the training data subset selection problem for these classifiers turns out to be a constrained submodular maximization problem. 

\subsection{Our Contributions}
In this paper, we  present a unified framework for data subset selection using two models for data summarization. The first is Facility Location (which models representation) and the second is  Minimum Dispersion (which models diversity). We argue for the utility of these functions and intuitively highlight the cases in which one of the models would work better compared to the other. We subsequently provide four concrete use-cases of our framework. We  divide our applications into supervised data selection (where you know the labels), unsupervised data selection (where we have no label information) and active learning.
\begin{enumerate}
\item \emph{Application 1: Supervised Data Subset Selection for Quick Training/Inference:} This application investigates the use of data subset selection for quick model training and inference. As an example, we look at KNN classification with features extracted from CNNs. This is particularly relevant here since complexity of inference of this non-parametric classifier is directly proportional to the number of training examples.
\item \emph{Application 2: Supervised Data Subset Selection for Quick Hyper-parameter tuning:} Another application of data subset selection is  selection of a representative yet smaller subset for hyper-parameter tuning for faster turn-around time. With this smaller subset (say, for example 5\% of the data), we can run several quicker experiments to find the optimal hyper-parameters. Once the hyper-parameters have been tuned, we can then train it on the full data.
\item \emph{Application 3: Unsupervised Data Subset Selection for Labeling from Video Data:} Often we need to custom train models on specific scenarios (e.g. data from self driving cars), and often this data comes from videos. We can use unsupervised data summarization to get a representative set of frames (from, say, a large video) which can then be labeled to create a training dataset. The role of diversity is clearly evident here since videos tend to have a lot of redundancy in them.
\item \emph{Application 4: Diversified Active Learning:} Lastly, we use our submodular data subset selection for diversified active learning, wherein we combine active learning (for example, uncertainty sampling) with diversified selection.
\end{enumerate}
We provide insights into the choice of two different summarization models for subset selection. The models have very diverse characteristics, and we try to argue scenarios where we might want to use one over the other.  We empirically demonstrate the utility of our framework and all the applications above across the board for several complex computer vision tasks including Gender Recognition, Object Recognition, Object Detection, Face Recognition and Scene Recognition. We show how using the right submodular models can provide as much as 5 - 10\% improvements over existing baselines in each of these four applications. We also point out here that the techniques proposed here are orthogonal to related work on transfer Learning, Zero shot learning and self supervised learning. As an example, one of the flavors we consider for diversified active learning, is where we don't train an end-to-end CNN but use transfer learning for training the models. It will be interesting to study how this work can be extended to other flavors including zero shot learning, semi-supervised learning etc.
%\begin{enumerate}
%\item We demonstrate the utility of subset selection in training deep models with a much smaller subset of training data, yet without significant compromise in accuracy for a variety of computer vision tasks. This is also particularly relevant for training a nearest neighbor model given that the complexity of inference of this non-parametric classifier is directly proportional to the number of training examples
%\item We demonstrate the utility of diversified subset selection in reducing labeling effort by eliminating the redundant instances amongst those chosen to be labeled by the traditional uncertainty method in every iteration of mini-batch adaptive active learning. 
%\item We provide insights to the choice of two different diversity models for subset selection, namely Facility-Location (which models representation) and Dispersion Functions (which models diversity). Both functions have very different characteristics, and we try to argue scenarios where we might want to use one over the other. 
%\item We empirically demonstrate the utility of this framework for several complex computer vision tasks including Gender Recognition, Object Recognition and Scene Recognition. Moreover, we apply this to Fine tuning Deep Models (end to end experiments) as well as Transfer Learning. 
%\end{enumerate}
\section{Data Subset Selection and Active Learning}
\subsection{Data Subset Selection} \label{submod-dss}
Given a set $V = \{1, 2, 3, \cdots, n\}$ of items which we also call the \emph{Ground Set}, define a utility function (set function) $f:2^V \rightarrow \mathbf{R}$, which measures how good a subset $X \subseteq V$ is. The goal is then to have a subset $X$ which maximizes $f$ with a constraint that the size of the set is less than or equal to $k$. It is easy to see that maximizing a generic set function becomes computationally infeasible as $V$ grows. 
\begin{align}
\mbox{Problem 1: } \max\{f(X) \mbox{ such that } |X| \leq k\}
\end{align}

A special class of set functions, called submodular functions ~\cite{nemhauser1978analysis}, however, makes this optimization easy. Submodular functions exhibit a property that intuitively formalizes the idea of ``diminishing returns''. That is, adding some instance $x$ to the set $A$ provides more gain in terms of the target function than adding $x$ to a larger set $A'$, where $A \subseteq A'$.  Informally, since $A'$ is a superset of $A$ and already contains more information, adding $x$ will not help as much. Using a greedy algorithm to optimize a submodular function (for selecting a subset) gives a lower-bound performance guarantee of a factor of $1 - 1/e$ of optimal ~\cite{nemhauser1978analysis} to Problem 1, and in practice these greedy solutions are often within a factor of 0.98 of the optimal ~\cite{krause2008optimizing}. This makes it advantageous to formulate (or approximate) the objective function for data selection as a submodular function. 

Several diversity and coverage functions are submodular, since they satisfy this diminishing returns property. The ground-set $V$ and the items $\{1, 2, \cdots, n\}$ depend on the choice of the task at hand. Submodular functions have been used for several summarization tasks including Image summarization~\cite{tschiatschek2014learning}, video summarization~\cite{gygli2015video}, document summarization~\cite{lin2011class}, training data summarization and active learning~\cite{wei2015submodularity} etc.

Building on these lines, in this work we demonstrate the utility of subset selection in allowing us to train machine learning models using a subset of training data without significant loss in accuracy. In particular, we focus on \emph{Facility-Location} function, which models the notion of representativeness and the \emph{Dispersion} function, which models the notion of diversity.

\paragraph{Representation: } Representation based functions attempt to directly model representation, in that they try to find a representative subset of items, akin to centroids and medoids in clustering. The Facility-Location function ~\cite{mirchandani1990discrete} is closely related to k-medoid clustering. Denote $s_{ij}$ as the similarity between images $i$ and $j$. We can then define $f(X) = \sum_{i \in V} \max_{j \in X} s_{ij}$. For each image $i$, we compute the representative from $X$ which is closest to $i$ and add the similarities for all images. Note that this function, requires computing a $O(n^2)$ similarity function. However, as shown in~\cite{wei2014fast}, we can approximate this with a nearest neighbor graph, which will require much less storage, and also can run much faster for large ground set sizes.

\paragraph{Diversity Models: } Diversity based functions attempt to obtain a diverse set of keypoints. The goal is to have minimum similarity across elements in the chosen subset by maximizing minimum pairwise distance between elements. There is a subtle difference between the notion of diversity and the notion of representativeness. While diversity \emph{only} looks at the elements in the chosen subset, representativeness also worries about their similarity with the remaining elements in the superset. Denote $d_{ij}$ as a distance measure between images $i$ and $j$. Define a set function $f(X) = \min_{i, j \in X} d_{ij}$. This function, called Dispersion function is not submodular, but can be still be efficiently optimized via a greedy algorithm~\cite{dasgupta2013summarization}. A common choice of diversity models used in literature are determinantal point processes~\cite{kulesza2012determinantal}, defined as $p(X) = \mbox{Det}(S_X)$ where $S$ is a similarity kernel matrix, and $S_X$ denotes the rows and columns instantiated with elements in $X$. It turns out that $f(X) = \log p(X)$ is submodular, and hence can be efficiently optimized via the Greedy algorithm. However, unlike the Dispersion functions, this requires computing the determinant and is $O(n^3)$ where $n$ is the size of the ground set. This function is not computationally feasible for large scale and hence we do not consider it in our experiments. It is easy to see that maximizing the Dispersion function involves obtaining a subset with maximal minimum pairwise distance, thereby ensuring a diverse subset of snippets or key-frames. The Dispersion function is called Disparity-Min function (we shall use both inter-changeably in this paper).

\paragraph{Optimization Algorithms: } For cardinality constrained maximization (Problem 1), a simple greedy algorithm provides a near optimal solution. Starting with $X^0 = \emptyset$, we sequentially update $X^{t+1} = X^t \cup \mbox{argmax}_{j \in V \backslash X^t} f(j | X^t)$, where $f(j | X) = f(X \cup j) - f(X)$ is the gain of adding element $j$ to set $X$. We run this till $t = k$ and $|X^t| = k$, where $k$ is the budget constraint. It is easy to see that the complexity of the greedy algorithm is $O(nkT_f)$ where $T_f$ is the complexity of evaluating the gain $f(j | X)$. This simple greedy algorithm can be significantly optimized via a lazy greedy algorithm~\cite{minoux1978accelerated}. The idea is that instead of recomputing $f(j | X^t), \forall j \notin ^t$, we maintain a priority queue of sorted gains $\rho(j), \forall j \in V$. Initially $\rho(j)$ is set to $f(j), \forall j \in V$. The algorithm selects an element $j \notin X^t$, if $\rho(j) \geq f(j | X^t)$, we add $j$ to $X^t$ (thanks to submodularity). If $\rho(j) \leq f(j | X^t)$, we update $\rho(j)$ to $f(j | X^t)$ and re-sort the priority queue. The complexity of this algorithm is roughly  $O(k n_R T_f)$, where $n_R$ is the average number of re-sorts in each iteration. Note that $n_R \leq n$, while in practice, it is a constant thus offering almost a factor $n$ speedup compared to the simple greedy algorithm. One of the parameters in the lazy greedy algorithms is $T_f$, which involves evaluating $f(X \cup j) - f(X)$. One option is to do a na\"{\i}ve implementation of computing $f(X \cup j)$ and then $f(X)$ and take the difference. However, due to the greedy nature of algorithms, we can use memoization and maintain a precompute statistics $p_f(X)$ at a set $X$, using which the gain can be evaluated much more efficiently. At every iteration, we evaluate $f(j | X)$ using $p_f(X)$, which we call $f(j | X, p_f)$.  We then update $p_f(X \cup j)$ after adding element $j$ to $X$. Table 1 provides the precompute statistics, as well as the computational gain for the Facility Location and Dispersion Functions. In particular, it is easy to see that evaluating $f(j | X)$ na\"{\i}vely is much more expensive than evaluating $f(j | X, p_X)$. The following theorem provides the approximation guarantees for the greedy algorithm for the Facility Location and the Dispersion Functions:
\begin{theorem}~\cite{nemhauser1978analysis,dasgupta2013summarization}
The greedy algorithm is guaranteed to obtain an approximation guarantee of $1 - 1/e$ for Problem 1 when $f$ is the Facility Location function. Similarly, the greedy algorithm achieves an approximation factor of $1/2$ when $f$ is the Dispersion function. When $f$ is a linear combination of the Facility Location and Dispersion functions, we obtain an approximation factor of $1/4$.
\end{theorem}
\begin{table*}
\begin{center}
 \begin{tabular}{|| c | c |  c | c | c ||} 
 \hline
 $f(X)$ & $p_f(X)$ & $f(j | X, p_f)$ & $C_o$ & $C_p$\\ [0.5ex] 
 \hline\hline
 $\sum_{i \in V} \max_{k \in X} s_{ik}$ & $[\max_{k \in X} s_{ik}, i \in V]$ & $\sum_{i \in V} \max(p_f^i(X), s_{ij}) - p_f^i(X)$ & $O(n^2)$ & $O(n)$\\ 
 \hline
  $\min_{k,l  \in X, k \neq l} d_{kl}$ & $\min_{k, l \in X, k \neq l} d_{kl}$ & $\min\{p_f(X), \min_{k \in X} d_{kj}\} - p_f(X)$ & $O(|X|^2)$ & $O(|X|)$\\
 \hline
 \end{tabular}
\caption{List of the precompute statistics $p_f(X)$, gain evaluated using the precomputed statistics $p_f(X)$ and finally $C_o$ as the cost of evaluation the function without memoization and $C_p$ as the cost with memoization for Facility Location and Dispersion Functions. It is easy to see that memoization saves an order of magnitude in computation.}
\end{center}
\end{table*}

\subsection{Techniques for active learning}
Active learning can be implemented in three flavors ~\cite{settles2010active}. The first is Batch active learning - there is one round of data selection and the data points are chosen to be labeled without  any  knowledge  of  the  resulting  labels  that  will be returned. The second is Adaptive active learning - there are many rounds of data selection, each of which selects one data point whose label may be used to select the data point at future rounds. The third flavor is Mini-batch adaptive active learning - a hybrid scheme where in each round a mini-batch of data points are selected to be labeled, and that may inform the determination of future mini-batches. In our work, we shall focus on the mini-batch active learning scheme. As far as the query framework in active learning is concerned, the simplest and most commonly used query framework is uncertainty sampling ~\cite{lewis1994sequential} wherein an active learner queries the instances about which it is least certain how to label. Getting a label of such an instance from the oracle would be the most informative for the model. Another,  more theoretically-motivated query selection framework is the query-by-committee (QBC) algorithm ~\cite{seung1992query}. In this approach, a committee of models is maintained which are all trained on the current labeled set but represent competing hypothesis. Each committee member is then allowed to vote on the labeling of query candidates.  The most informative query is considered to be the instance about which they most disagree. There are other frameworks like Expected Model Change ~\cite{settles2008multiple}, Expected Error Reduction ~\cite{roy2001toward}, Variance Reduction ~\cite{cohn1994neural} and Density-Weighted methods ~\cite{settles2008analysis}. In this paper, we shall use uncertainty sampling as the active learning algorithm.

There are three common ways to compute uncertainty for each unlabeled data instance. 
\begin{align}
          u &= 1-\max_i p_i && \\
          u &= 1-(\max_i p_i- \max_{j \in \{C\}-\arg\max_i  p_i} p_j )\\
          u &= -\sum_i {p_i * \log_2 ( p_i )}
\end{align}
where $u = \text{uncertainty}$, $p_i = \text{probability of class $i$}$ and $\{C\} = \text{set of classes}\\$. The goal of uncertainty sampling is to pick a batch of unlabeled data instances with maximum uncertainty. 
%------------------------------------------------------------------------
%------------------------------------------------------------------------
\begin{figure*}
\begin{center}
\includegraphics[width = 0.45\textwidth]{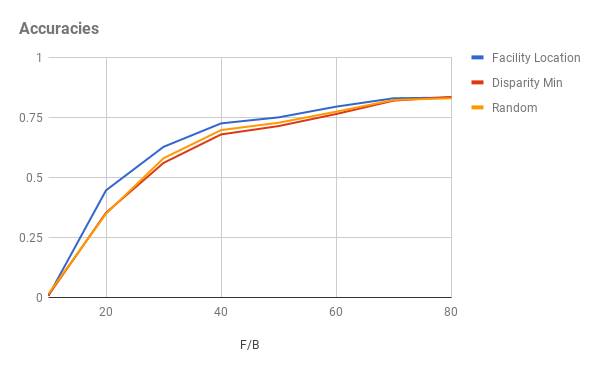}
~
\includegraphics[width = 0.45\textwidth]{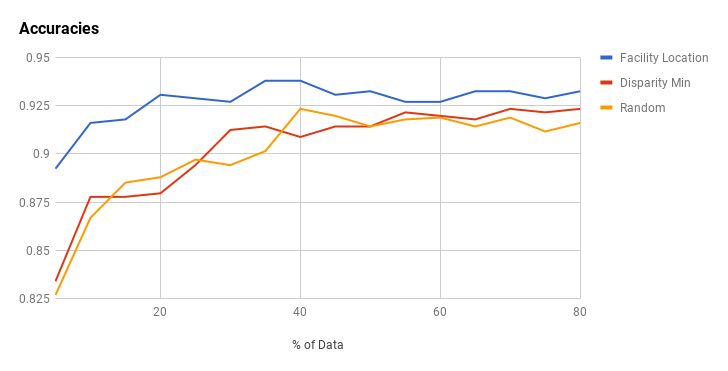}
~
\includegraphics[width = 0.45\textwidth]{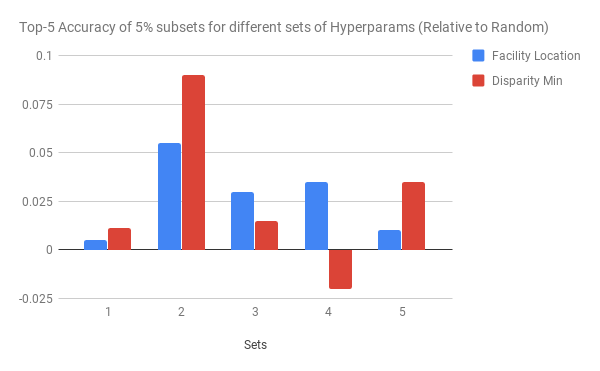}
~
\includegraphics[width = 0.45\textwidth]{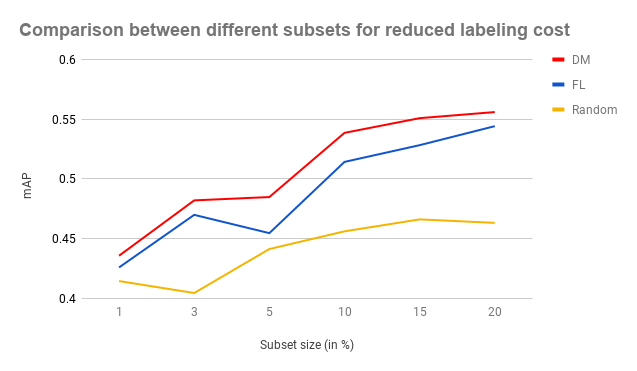}
\end{center}
\caption{Blue (Facility Location), Red (Disparity Min) and where-ever applicable, Orange (Random). The top two figures show the results for application 1 (KNN Classification) on the Face and Gender Datasets. The Bottom left plot shows the results for application 2 (hyper-parameter tuning) for 5\% of the subset for various sets of hyper-parameters (ImageNet) and finally the Bottom right plot shows the results for application 3 on the Video dataset (Video Surveillance Object).}
\label{fig:app1-3}
\end{figure*}

\begin{figure*}
\begin{center}
\includegraphics[width = 0.48\textwidth]{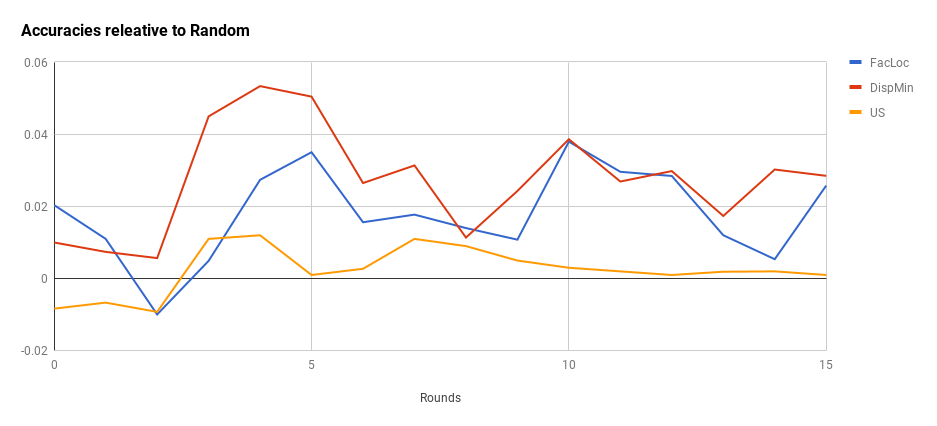}
~
\includegraphics[width = 0.48\textwidth]{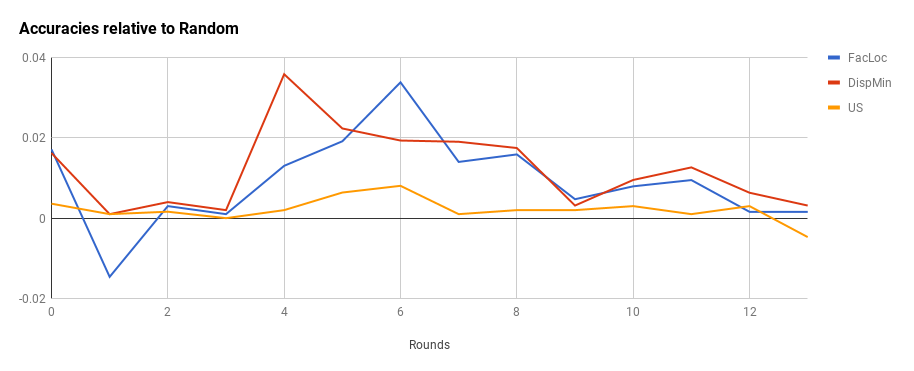}
~
\includegraphics[width = 0.48\textwidth]{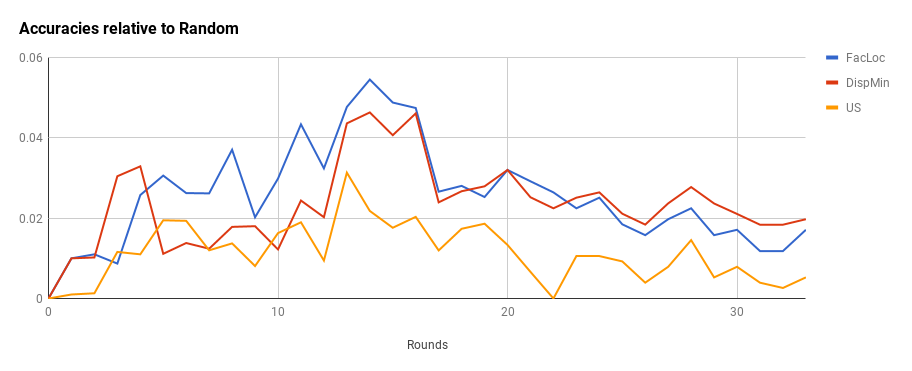}
~
\includegraphics[width = 0.48\textwidth]{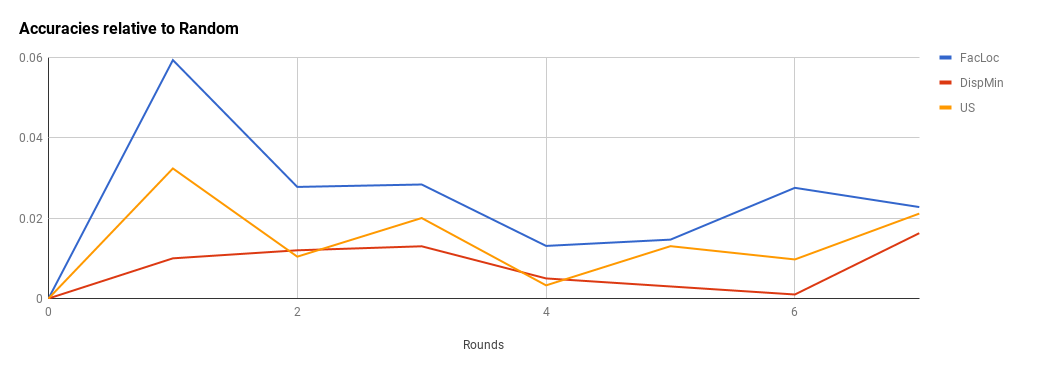}
~
\includegraphics[width = 0.48\textwidth]{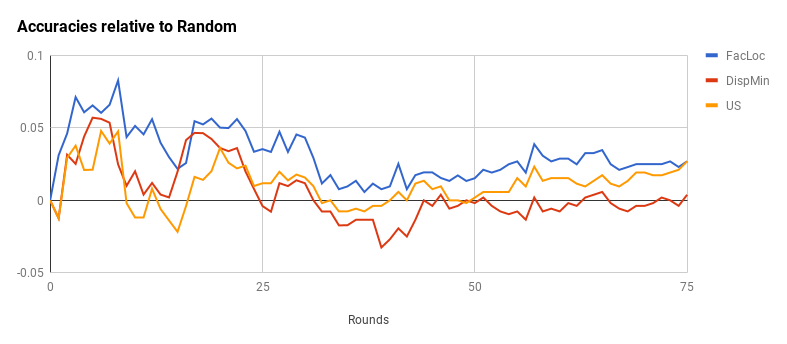}
~
\includegraphics[width = 0.48\textwidth]{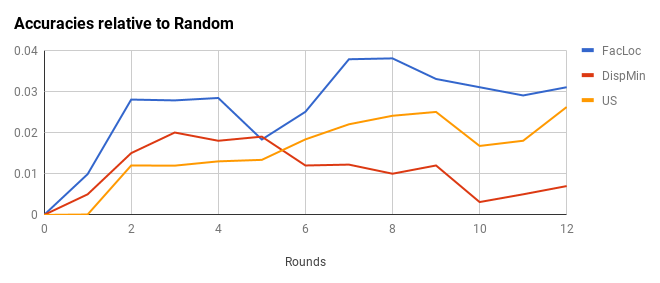}
\end{center}
\caption{Results for application 4: Active Learning. Accuracies relative to Random: Blue (Facility Location), Red (Dispersion) and Orange (Uncertainty Sampling). Top Left: Adience, Top Right: Cats vs Dogs, Middle Left: Face, Middle Right: ImageNet, Bottom Left: Caltech-101, and Bottom Right: MIT-67. The plots on the top row are obtained using finetuning, while the rest (middle and bottom rows) are using transfer learning.}
\label{fig:app4}
\end{figure*}

\section{Applications of our Framework}
In this section, we describe the applications of our Data Subset Selection Framework from Section~\ref{submod-dss}. We empirically establish the utility of this framework on four different data selection and active learning applications 

\subsection{Supervised DSS for Quick Training/Inference}
As a first application, we evaluate the efficiency of our subset selection methods to enable learning from lesser data, yet incurring minimal loss in performance. Towards this, we apply Facility-Location and Dispersion model based subset selection techniques for training a nearest neighbor classifier (kNN). For a given dataset we run several experiments to evaluate the accuracy of kNN, each time trained on a different sized subset of training data. We take subsets of sizes from 5\% to 100\% of full training data, with a step size of 5\%. For each experiment we report the accuracy of kNN over the hold-out data. We also compare these results against kNN trained on randomly selected subsets. Concretely, denote $f$ as the measure of \emph{information} as to how well a subset $X$ of training data performs as a proxy to the entire dataset. We then model this as an instance of Problem 1 with $f$ being the Facility-Location and Dispersion Functions. Since this is a supervised data subset selection problem, we can also use the label information. We partition the ground set $V$ as $\{V_1, \cdots, V_k\}$ and $k = |\mathcal C|$. Given a submodular function $f$, define the label-aware version of $f$ as,
\begin{align}
f^y(X) = \sum_{i = 1}^k f(X \cap V_i)
\end{align}

\subsection{Supervised DSS for Hyper-Parameter Tuning}
Another application of our Data subset selection framework is to select a representative and diverse subset for quick hyper-parameter training. Note that this is an application of supervised data selection, since we know the labels here. We use the Facility Location and Min Dispersion functions to get a subset of the data, and use this subset for tuning typer-parameters. Note that the subset is typically 5 - 10\% of the entire data, and hence it is much easier to tune the hyper-parameters on this subset. Once the hyper-parameters are tuned, we train the model on the entire data using the tuned hyper-parameters. We expect our summarization models to perform better compared to simple random sampling, and thereby be more representative for hyper-parameter tuning. 

\subsection{Data Subset Selection on Massive Datasets for Labeling}
As another application, we consider the problem of labeling massive datasets, specifically when the data comes from videos. As an example, consider the applications discussed in~\cite{dubal2018deployment}, where they observe that model customization can substantially improve performance for image classification and object detection tasks. Moreover, often the data here comes from videos where naturally there is a lot of redundancy. Unlike the supervised data subset selection discussed above, the data here is unlabeled, so we need to apply unsupervised data subset selection here. We introduce a surveillance dataset comprising of around 20 videos from various scenarios (indoor, coartyards, outdoor scenario like roads, traffic etc.) with frames sampled at 1 FPS. 

\subsection{Diversified Active Learning}
Finally, we study the efficiency of our subset selection methods in the context of mini-batch adaptive active learning to demonstrate savings on labeling effort. We propose a Submodular Active Learning algorithm. The idea is to use uncertainty sampling as a filtering step to remove less informative training instances. After filtering, we obtain a subset $F$. Denote $\beta$ as the size of the filtered set $F$. We then select a subset $X$ by solving Problem 1.
\begin{align}
\max\{f(X) \mbox{ such that } |X| \leq B, X \subseteq F\}
\end{align}
where $B$ is the number of instances selected by the batch active learning algorithm at every iteration. In our experiments, we use $f$ as Facility-Location and Disparity-Min functions. For a rigorous analysis and a fair evaluation, we implement this both in the context of transfer learning - as well as fine tuning. For transfer learning, we extract the features from a pre-trained CNN relevant to the computer vision task at hand and train a Logistic Regression classifier using those features. 

The basic diversified active learning algorithm is as mentioned in the Algorithm \ref{Algo:Goal-2} 
\begin{algorithm}[H]
\caption{Goal-2: Submodular Active Learning}\label{Algo:Goal-2}
\begin{algorithmic}[1]
\STATE Start with a small initial seed labeled data, say $L$
\FOR{each round 1 to $T$ }
\STATE Fine tune a pre-trained CNN (Goal-2a) or train a Logistic Regression classifier (Goal-2b) with the labeled set $L$
\STATE Report the accuracy of this model over the hold-out data.
\STATE Using this model, compute uncertainties of remaining unlabeled data points $U$, and select a subset $F, F \subseteq U$ of the most uncertain data points. 
\STATE Solve Problem 2 with Facility-Location and Disparity-Min Functions 
\STATE Label the selected subset and add to labeled set $L$
\ENDFOR
\end{algorithmic}
\end{algorithm}

\begin{table*}[ht]
\centering
\begin{adjustbox}{width=1\textwidth}
\small
\begin{tabular}{rlrrrrrrr}
  \hline
 & Name & NumClasses & NumTrain & NumTrainPerClass & NumHoldOut & NumHoldOutPerClass \\ 
  \hline
  &Adience ~\cite{Adience, eidinger2014age} & 2 & 1614 & 790(F), 824(M) & 807 &  395(F), 412(M)   \\
  &Caltech-101  ~\cite{Caltech-101, fei2007learning} & 101 & 7900 & 40-56 & 1246 &  8-15 \\
    &MIT-67 ~\cite{MIT67, quattoni2009recognizing} & 67 & 10438 & 68-490 & 5159 &  32-244  \\
  &CatsVsDogs ~\cite{CatsvsDogs, elson2007asirra} & 2 & 16668 & 8334 & 8332 &  4166 \\
   & ImageNet~\cite{russakovsky2015imagenet} & 1000 & ~1.2M & 1000 & 50000 & 50 \\
  & Video Surveillance Object & 8 & 76300 & ~5000 - 12000 & 10,211 & ~500 - 3000 \\
    &GenderData & 2 & 2200 & 1087(F), 1113(M) & 548 &  249(F), 299(M)  \\
  &FaceData & 255 & 2345 & 7-8 & 552 &  2-4  \\
   \hline
\end{tabular}
\end{adjustbox}
\caption{Details of datasets used in our experiments} 
\label{tab:datasets}
\end{table*} 
\normalsize

The parameters $B$ and $\beta$ of FASS: $B$ represents the percentge of images that are to be labeled at the end of each round and added to the training set. $\beta$ specifies what percentage of data (sorted in decreasing order of their hypothesized uncertainties by the current model) forms the ground set (for subset selection) in every round. 

While selecting $\beta$\% most uncertain samples at each round, any other data point having the exact same uncertainty as the last element of this set is also added to the uncertainty ground set.

\begin{table*}[ht]
\centering
\begin{adjustbox}{width=1\textwidth}
\small
\begin{tabular}{rlrrrrrrr}
  \hline
 & Task & Dataset & Model & Parameters \\ 
  \hline
  & Gender Recognition & Adience & VGGFace/CelebFaces ~\cite{VggFace_CelebFaces} & $B=0.9\%,\beta=10\%$\\
  & Object Recognition & Cats vs Dogs & GoogleNet/ImageNet ~\cite{GoogleNet_ImageNet} & $B=1\%,\beta=10\%$\\
 & Gender Recognition & FaceData & VGGFace/CelebFaces ~\cite{VggFace_CelebFaces} & $B=0.5\%,\beta=10\%$\\
  & Scene Recognition  & MIT-67 & GoogleNet/Places205 ~\cite{GoogleNet_Places} & $B=2\%,\beta=10\%$\\
  & Object Recognition & Caltech-101 & VGGFace/CelebFaces ~\cite{VggFace_CelebFaces} & $B=1\%,\beta=10\%$\\
  & Object Recognition & ImageNet & GoogleNet/ImageNet ~\cite{GoogleNet_ImageNet} & $B=1\%,\beta=10\%$\\
  \hline
\end{tabular}
\end{adjustbox}
\caption{Experimental setup for application 4} 
\label{tab:experiments}
\end{table*} 
\normalsize
\section{Experiments and Datasets}
In Table \ref{tab:datasets} we present the details of different datasets used by us in above experiments along with the train-validate split for each. Out of the 8 datasets used, 5 datasets are publicly available datasets and 3 of the datasets are our custom datasets (FaceData, GenderData and Video Surveillance Object).

For application 1, we evaluate kNN with k=5, and in application 2, we perform hyper-parameter tuning on ImageNet with a subset of 5\% of the data.
With application 3, we use our custom video dataset consisting of 76300 Images for Object detection with the following classes: Person, Car, Bus, motorbike, Bicycle and Three-wheeler. Our dataset comprises of videos from roads (outdoor), coartyard (outdoor), office (indoor). Finally, For application 4 experiments, the number of rounds was experimentally obtained for each experiment as the percentage of data that is required for the model to reach saturation. The values of $B$ and $\beta$ were empirically arrived at and are mentioned in Table \ref{tab:experiments} along with the details about the particular Computer Vision task and dataset used for each set of experiments. 

In Table\ref{tab:experiments}, the ``Model'' column refers to the pre-trained CNN used to extract features and/or for finetuning. For example, ``GoogleNet/ImageNet'' refers to the GoogleNet model pre-trained on ImageNet data. In the ``Parameters'' column, $B$ and $\beta$ refer to the parameters in FASS and are mentioned in percentages. We use Caffe deep learning framework ~\cite{jia2014caffe} for all our Image classification experiments and DarkNet~\cite{redmon2016you} for Object Detection.

\section{Results}
We present the resules for application 1 in the top two plots in Figure \ref{fig:app1-3}. As hypothesized, using only a portion of training data (about 40\% in our case) we get similar accuracy as with using 100\% of the data. Moreover, Facility Location performs much better than random sampling and Disparity Min, proving that Facility Location function is a good proxy for nearest neighbor classifier. Moreover, we see this phenomenon on both Gender Recognition and Face Recognition tasks. Also note that this aligns with the theoretical results shown in~\cite{wei2015submodularity}.

For application 2, we consider the ImageNet~\cite{russakovsky2015imagenet}. We choose a subset of  5\% of the dataset through supervised data subset selection obtained via random sampling, Disparity Min and Facility Location. We use SGD with Momentum as the learning algorithm and tune the learning rate and momentum parameters. In particular, we select five sets of parameters for the learning rate $\alpha$ and momentum $\mu$ as: $(\alpha, \mu) = (0.01, 0.9), (0.02, 0.8), (0.015, 0.85), (0.005, 0.95)$, and $(0.001, 0.99)$. Figure~\ref{fig:app1-3} bottom left shows the results of the 5\% subsets obtained via Disparity-Min and Facility Location relative to Random. We notice that the Facility Location function generally has a positive gain compared to random, and except for one of the hyper-parameters disparity-min also beats random. We next select the hyper-parameters for random, disparity min and Facility Location which obtain the best accuracy on the validation set (which was Set 1, Set 5 and Set 4 respectively). Using this hyper-parameters, we train imagenet using the complete dataset. We observe that the hyper-parameters chosen by FL obtain the best result (around 10.5\% improvement in Top-5 accuracy compared to the hyper-parameters chosen by the random subset). The hyper-parameters chosen by Disparity-Min achieve around 52.5\% top-5 accuracy, while the ones with Facility Location achieve close to 59.5\%. In comparison the random hyper-parameters achieve a top-5 accuracy of 42\%. All these results are using the GoogleNet architechture.

%0.5\% accuracy improvements compared to the random hyper-parameters. This shows that selecting better subsets for hyper-parameter tuning, can result in good choices of hyper-parameters since these subsets tend to be more representative of the entire dataset. 

In the case of application 3, we perform unsupervised data subset selection. The task here is to obtain a subset of images from a larger dataset where the frames are taken from videos for labeling. We consider the problem of Object Detection using YOLOv2~\cite{redmon2016you}. We achieve subsets of various sizes using unsupervised video summarization and compare the results to random. The results are shown in Bottom right in Figure~\ref{fig:app1-3}. We see that Disparity Min has the best results followed closely by Facility Location. Both these methods achieve almost a 5\% improvement compared to random subset in mAP. This is expected since Disparity Min tends to pick a diverse set of images for training, thereby ensuring a good mix between the classes for training, as compared to random which is not aware of the redundancy in the images. The fact that Disparity-Min models diversity also suggests that diversity is more important here compared to representation.

Finally, we compare the different models for application 4. First, we notice that almost always, the subset selection techniques and uncertainty sampling always outperform random sampling. Moreover, different subset selection techniques work well for different problem contexts, with the intuition behind these results described below.

We present the results for application 4 in Figure \ref{fig:app4}.  Recall that Disparity-Min selects most diverse elements, while Facility Location picks the representative items. Figure \ref{fig:app4} (top row) demonstrates the results on Adience and Cats vs Dogs datasets. We notice that Diversity works better than representation in both these cases, particularly when the size of the dataset is small. This is because both these problems are two class classification, and there is a lot of similarity in the dataset (for example, there is more than one image for the same person, or similar looking cats or dogs). Therefore it is not surprising that Diversity works really well in these settings. Next we compare the results on Object recognition, face recognition and scene recognition datasets (Middle and Bottom rows of Figure~\ref{fig:app4}): ImageNet, Caltech-101, MIT 67 and Face dataset. In these datasets, we have several classes and there is not a lot of redunduncy. Disparity Min tends to pick the outlier images which often does not make sense here. In this case, we see that the representation model (Facility Location) works the best. In each case, we see that subset selection algorithms on top of Uncertainty Sampling outperform vanilla Uncertainty Sampling and Random Selection. Moreover, the effect of diversity and representation reduce as the number of unlabeled instances reduce in which case Uncertainty sampling already works well. In practice, that would be the point where the accuracy saturates as well.

\section{Conclusions and Lessons Learnt}
This paper demonstrates the utility of subset selection in training models for a variety of standard Computer Vision tasks. Subset selection functions like Facility-Location and Disparity-Min naturally capture the notion of representativeness and diversity which thus help to eliminate redundancies in data. We show the practical utility of this data subset selection in four applications. The goal of the first application is to use data subset selection for reducing training and inference time. We demonstrate this for KNN classification and show that Facility Location (representation models) perform the best with considerable improvement over a random sampling. Next, we look at an application of hyper-parameter tuning for Image classification. We demonstrate this on ImageNet and show that the subset achieved by Facility Location is represents the entire dataset in a better way compared to a random subset. We also show that the best hyper-parameters on tunings performed over the representative subset achieves a better accuracy compared to the random subset. We also see, as expected, that the models trained on the Facility Location subset consistently outperform the models trained on random subsets. We then study an application of video data summarization for object detection. We see here that the diversity measure makes more sense since there is naturally a lot of redundancy in video datasets. We see consistently that the models trained on diverse subset beats the random subset by almost 5\% in mAP. Finally, we demonstrate the benefit of data summarization with uncertainty sampling. We see that Disparity works best when there is the need for diverse selection (when the dataset tends to have a higher amount of redundancy) while the Facility Location model works best otherwise. In either case, we see that both these models consistently outperform uncertainty sampling alone, which itself performs generally better than random.

\section{Acknowledgements}
The authors would like to thank Suyash Shetty, Anurag Sahoo, Narsimha Raju and Pankaj Singh for discussions and useful suggestions on the manuscripts.
%We leverage this and show its practical utility in two settings - ability to learn from less data (Goal-1) and reduced labeling effort (Goal-2). As discussed, these subset selection methods lend themselves well to optimization and hence are easy to implement using simple greedy algorithms. Our results also enabled us to develop insights on the choice of subset selection methods depending on a particular setting. As noted earlier, when ground set is diverse to begin with - either because of diversity \emph{within} a class or diversity due to more number of classes - both Disparity-Min and Facility-Location perform well. On the other hand, when data points within a class are similar, Disparity-Min, in want of picking most mutually diverse images, ends up picking outliers (with respect to items in a class) and confuses the model. We hope that the intuitions obtained from this paper will help practioners develop active learning and subset selection techniques for image recognition tasks. As another potential application, in addition to enabling learning from less data and helping reduce labeling costs, such subset selection techniques can well complement the existing annotation frameworks like ~\cite{vondrick2013efficiently, schoeffmann2015video}.

\bibliographystyle{ieee}
\bibliography{bmvc}

\end{document}